\documentclass[runningheads]{llncs}

\usepackage[T1]{fontenc}
\usepackage{graphicx}
\usepackage{marvosym}
\usepackage{amssymb}
\usepackage{amsmath}
\usepackage{bm}
\usepackage{bbm}
\usepackage{booktabs}
\usepackage[colorlinks=true,linkcolor=blue,citecolor=blue,urlcolor=blue,]{hyperref}
\usepackage{wrapfig}

\begin{document}

\title{CausalCLIPSeg: Unlocking CLIP's Potential in Referring Medical Image Segmentation with Causal Intervention}

\titlerunning{CausalCLIPSeg}
\author{Yaxiong Chen\inst{1,2} \and 
Minghong Wei\inst{1}\thanks{Work done during an internship at MedAI Technology (Wuxi) Co. Ltd.} \and
Zixuan Zheng\inst{3} \and
Jingliang Hu\inst{3} \and
Yilei Shi\inst{3} \and
Shengwu Xiong\inst{1,2} \and
Xiao Xiang Zhu\inst{4} \and
Lichao Mou\inst{3}\textsuperscript{(\Letter)}}


\authorrunning{Y. Chen et al.}

\institute{Wuhan University of Technology, Wuhan, China \and Shanghai Artificial Intelligence Laboratory, Shanghai, China \and MedAI Technology (Wuxi) Co. Ltd., Wuxi, China\\\email{lichao.mou@medimagingai.com} \and Technical University of Munich, Munich, Germany}

\maketitle

\begin{abstract}
Referring medical image segmentation targets delineating lesions indicated by textual descriptions. Aligning visual and textual cues is challenging due to their distinct data properties. Inspired by large-scale pre-trained vision-language models, we propose CausalCLIPSeg, an end-to-end framework for referring medical image segmentation that leverages CLIP. Despite not being trained on medical data, we enforce CLIP's rich semantic space onto the medical domain by a tailored cross-modal decoding method to achieve text-to-pixel alignment. Furthermore, to mitigate confounding bias that may cause the model to learn spurious correlations instead of meaningful causal relationships, CausalCLIPSeg introduces a causal intervention module which self-annotates confounders and excavates causal features from inputs for segmentation judgments. We also devise an adversarial min-max game to optimize causal features while penalizing confounding ones. Extensive experiments demonstrate the state-of-the-art performance of our proposed method. Code is available at \url{https://github.com/WUTCM-Lab/CausalCLIPSeg}.

\end{abstract}

\section{Introduction}
Medical image segmentation is a crucial prerequisite for numerous clinical applications \cite{zhou2017fixed,gering1999integrated} such as computer-aided diagnosis, surgical planning, and image-guided intervention. However, the inherent complexity and variability of medical images often poses significant challenges for automatic segmentation methods. In particular, lesions can induce severe distortions and intensity alterations that confuse segmentation models. Incorporating prior knowledge, e.g., lesion count and location, may help.
\par
Medical notes written by physicians provide descriptive details and contextual information about medical images that can aid segmentation models in better delineating lesions. Moreover, accessing these texts incurs marginal additional cost, as they are already stored in electronic medical record systems within medical institutions. Unlike manual annotations which require extra labor, we can utilize the existing medical notes to improve segmentation performance without needing additional data collection and labeling. As a pioneering work, \cite{li2023lvit} extends the QaTa-COV19 dataset by adding text descriptions and designs a referring segmentation network. Yet so far, there is a paucity of literature focusing on this multi-modal segmentation task utilizing both images and texts.
\par
In the field of multi-modal deep learning, CLIP~\cite{radford2021learning}, a large-scale pre-trained vision-language model based on contrastive learning, has recently attracted increasing attention. CLIP, in contrast to uni-modal foundation models like SAM \cite{kirillov2023segment}, is better suited for multi-modal tasks. It learns image-level visual concepts from 400 million image-text pairs, enabling it to understand semantic relationships between images and texts. CLIP has proven highly effective across a diverse set of uni-modal and multi-modal downstream tasks~\cite{radford2021learning,ali2023clip,conde2021clip,wang2021actionclip,zhu2023learning,luo2022clip4clip,fang2021clip2video}, suggesting its potential utility for the challenging problem of referring medical image segmentation.
However, the rich knowledge encapsulated in CLIP is derived from its image-level training objective. Consequently, directly applying CLIP does not yield optimal performance on pixel-level dense prediction tasks.
In this paper, we employ CLIP for medical image segmentation from referring expressions, which has not been thoroughly investigated. Our key insight is that even though CLIP has not been trained on medical data, we can inherit much of its latent space's virtue by enforcing its powerful and semantic structure onto the medical domain.
\par
On the other hand, medical images often contain lesions with ambiguous boundaries~\cite{xie2022uncertainty,wang2021boundary,wang2023medical}, while some lesion-free regions have similar appearances to lesions, confusing segmentation models. This arises due to confounders that induce models to learn spurious correlations based on conventional likelihood estimation. Moreover, the use of CLIP exacerbates this problem to some extent, as CLIP's training objective differs from the task of referring medical image segmentation. Mitigating such confounding bias necessitates causal intervention~\cite{pearl2009causality}. For instance, collecting images of a specific object in all contexts causes models to focus on the target rather than context. However, exhaustively collecting medical images with all possible lesion backgrounds is infeasible. To address this, we propose a causal intervention module that self-annotates confounders in an unsupervised manner and excavates causal features from inputs, improving segmentation. We assume inputs comprise causal and non-causal factors, with only the former determining segmentation judgments. The module's goal is to suppress confounding features and retain causal ones. Additionally, we devise an adversarial min-max game, optimizing causal features while penalizing confounding ones. Our approach facilitates learning lesion-relevant representations without requiring exhaustive sampling of contexts. 
\par
Our technical contributions are three-fold: 
\begin{itemize}
    \item We introduce CLIP to referring medical image segmentation and design a framework called CausalCLIPSeg.
    \item We propose a causal intervention module to eliminate confounding bias.
    \item Our approach achieves state-of-the-art performance on the QaTa-COV19 dataset.
\end{itemize}

\subsubsection{Background of Causality}
In the context of this work, confounding bias refers to the interference of background factors and other extraneous variables in the model's segmentation of lesion areas. Spurious correlation is a related concept to confounding bias. A meaningful causal relationship is the ideal scenario where a model relies solely on visual features of lesion areas to generate lesion masks, without being influenced by confounding factors. Excavating causal features refers to extract visual features that satisfy the meaningful causal relationship. Self-annotating confounders refers to our model's ability to adaptively extract confounders.

\begin{figure*}[t]
\begin{center}
\includegraphics[width = 0.95\linewidth]{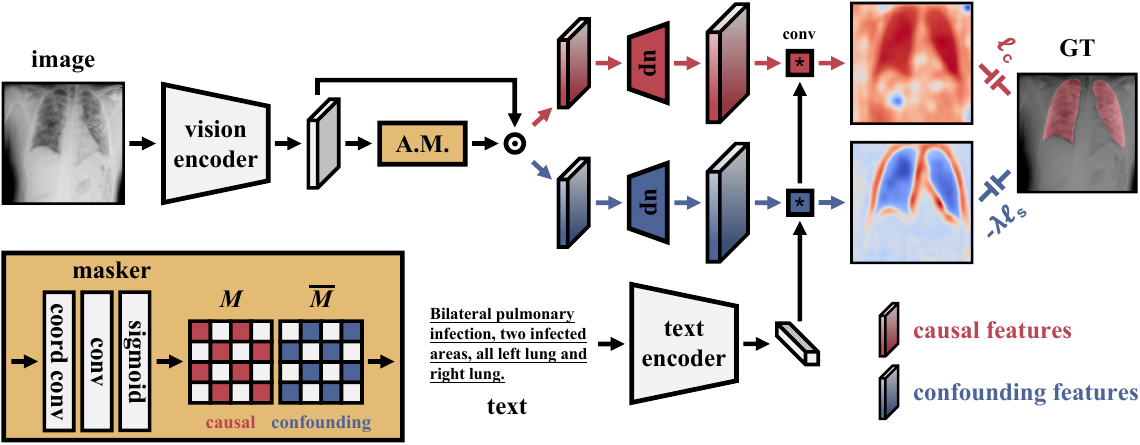}
\end{center}
\caption{Overview of the proposed CausalCLIPSeg model.}
\label{main}
\end{figure*}

\section{Method}

Fig.~\ref{main} illustrates the proposed CausalCLIPSeg model which comprises three key components: (i) CLIP-driven text and vision encoders that extract multimodal feature representations, (ii) a decoder that achieves text-pixel alignment, and (iii) a causal intervention module that eliminates confounding bias between segmentation targets and spurious correlations.

\subsection{CLIP-Driven Text and Vision Encoders}
\subsubsection{Text Encoder}
We employ the same modified Transformer architecture from~\cite{radford2021learning} as our text encoder. For an input expression, we tokenize it into a text sequence using a byte pair encoding (BPE) tokenizer with a 49,152 vocabulary size, identical to that used in GPT-2~\cite{radford2019language}. The text sequence is bracketed with $\texttt{[SOS]}$ and $\texttt{[EOS]}$ tokens. The Transformer performs on this sequence, and activations of its highest layer at the $\texttt{[EOS]}$ token are transformed as a global textual representation $\bm{\tau}\in\mathbb{R}^{T}$, where $T$ is the feature dimension.

\subsubsection{Vision Encoder}
For the vision encoder, we choose pre-trained ResNet-101. For an input image $\bm{I}\in\mathbb{R}^{1\times H\times W}$, we extract multi-scale visual features from stages 2 to 4, denoted as $\bm{F}_i$. Here, $H$ and $W$ represent the height and width of the image.

\subsection{Cross-Modal Decoding for Text-Pixel Alignment}
Our decoder $\mathcal{D}$ aligns the text feature $\bm{\tau}$ with pixels in the input image to generate a segmentation mask. Specifically, $\bm{\tau}$ is first mapped into a $D$-dimensional embedding space through a learnable linear projection, where $D=C\times K\times K+1$. The resulting embedding is further split into $\bm{W}_{\tau}\in\mathbb{R}^{C\times K\times K}$ and $b_{\tau}\in\mathbb{R}$. On the other hand, visual feature maps $\bm{F}$ are upsampled to the full resolution by interpolation and convolution. Then, the textual and visual features are cross-correlated:
\begin{equation}
\label{eq:decoder}
\bm{S}=\mathcal{D}(\bm{\tau},\bm{F})=\bm{W}_{\tau}\ast\mathrm{up}(\bm{F})+b_{\tau}\mathbbm{1} \,,
\end{equation}
where $b_{\tau}\mathbbm{1}$ denotes a signal which takes value $b_{\tau}\in\mathbb{R}$ in every location.
\par
Eq.~(\ref{eq:decoder}) amounts to performing an exhaustive cross-modal search of the input linguistic expression over the input image. The goal is for activations of the response map (left-hand side of Eq.~(\ref{eq:decoder})) to correspond to segmentation targets.

\subsection{Causal Intervention Module}

To enable the model to learn more robust semantic features, we introduce a causal intervention module that acts directly on the representations learned by the CLIP encoders.

\subsubsection{Causal View}

\begin{wrapfigure}{r}{4cm}
\vspace{-0.30in}
\centering
\includegraphics[width=0.22\textwidth]{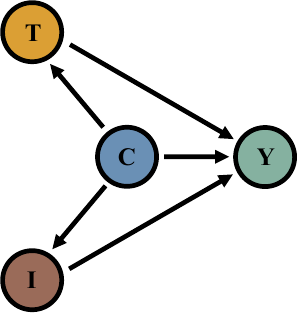}
\caption{Structural causal model for referring medical image segmentation.}
\label{causal}
\vspace{-0.20 in}
\end{wrapfigure}

We introduce the formulation of causality for referring medical image segmentation by using a structural causal model. Specifically, it is constructed by examining causalities among four key components: expression $\bm{T}$, image $\bm{I}$, ground truth mask $\bm{Y}$, and confounder $\bm{C}$. As illustrated in Fig.~\ref{causal}, each direct link denotes a causal relationship between two nodes. The link $\bm{I},\bm{T}\rightarrow \bm{Y}$ represents the desired causal effect from lesions in the image $\bm{I}$ to the segmentation mask $\bm{Y}$. We refer to a segmentation model as unbiased if it identifies $\bm{I},\bm{T}\rightarrow \bm{Y}$ based solely on lesion information. The path $\bm{I},\bm{T}\leftarrow \bm{C}\rightarrow \bm{Y}$ denotes that imaging conditions $\bm{C}$ determine pixel intensities in $\bm{I}$; for example, $\bm{C}$ encapsulates artifacts, acquisition protocols, and scanner models which influence image characteristics. Since models inevitably utilize imaging cues for segmentation, the link $\bm{C}\rightarrow \bm{Y}$ also exists. In this structural causal model, we clearly see how $\bm{C}$ confounds $\bm{I},\bm{T}$ and $\bm{Y}$ through the backdoor path $\bm{I},\bm{T}\leftarrow \bm{C}\rightarrow \bm{Y}$. For instance, while no causal link exists between background pixels and the segmentation mask, the backdoor path creates a spurious correlation between them (through $\bm{C}$), yielding incorrect lesion predictions.

\subsubsection{Adversarial Masking}
To be causally sufficient towards segmentation, we design an adversarial masking mechanism that iteratively detects features that contain relatively more causal information and separate them from confounding ones via a masker. Specifically, the masker produces a pair of complementary attention masks $\bm{M}$ and $\overline{\bm{M}}$ controlling every location in feature maps. $\bm{M}$ is for attending to features of the causal effect, while $\overline{\bm{M}}$ is for attending to the confounding effect. Note that these masks are generated during both training and inference, and they are learned and input-specific.
\par
We denote the causal and confounding features as $\bm{F}^c_i$ and $\bm{F}^s_i$, respectively. They can be obtained by our module as follows:
\begin{equation}
\begin{split}
\bm{M}_i&=\mathrm{sigmoid}(\mathcal{F}(\bm{F}_i)) \,, \\
\overline{\bm{M}}_i&=1-\bm{M}_i \,, \\
\bm{F}_i^c&=\bm{M}_i\odot\bm{F}_i \,, \\
\bm{F}_i^s&=\overline{\bm{M}}_i\odot\bm{F}_i \,,
\end{split}
\end{equation}
where $\mathcal{F}$ is implemented by cascading a coordinate convolution and a convolution, and $\odot$ represents element-wise product. 
By doing so, we disentangle $\bm{F}_i^c$ and $\bm{F}_i^s$ from the input features.
\par
Medical images often contain multi-scale lesions that necessitate model attention across scales. To obtain comprehensive representations, we fuse multi-scale visual features as follows. We first adopt the CARAFE operator~\cite{wang2019carafe} to transform causal and confounding features learned across stages into a unified scale. The scaled causal and confounding features are then concatenated separately. Finally, fused causal and confounding features, $\bm{F}^c$ and $\bm{F}^s$, are obtained via 1×1 convolutions over the concatenated features.

\subsubsection{Adversarial Training}
We optimize the causal intervention module along with the CLIP encoders and decoder with a min-max loss. We use two separate yet identical decoders $\mathcal{D}_c$ and $\mathcal{D}_s$ to predict two segmentation masks from $\bm{F}^c$ and $\bm{F}^s$, respectively. The loss is formulated as follows:
\begin{equation}
\mathcal{L}=\mathcal{L}_c-\lambda\mathcal{L}_s \,,
\end{equation}
where
\begin{equation}
\begin{split}
\mathcal{L}_c&=\ell_{ce}(\mathcal{D}_c(\bm{\tau},\bm{F}^c),\bm{Y}) \,, \\
\mathcal{L}_s&=\ell_{ce}(\mathcal{D}_s(\bm{\tau},\bm{F}^s),\bm{Y}) \,.
\end{split}
\end{equation}
Here, $\ell_{ce}$ denotes cross-entropy loss, $\bm{Y}$ is the ground truth mask of the input image, and $\lambda$ is a coefficient balancing the two loss terms. We optimize the entire model by minimizing $\mathcal{L}_c$ and maximizing $\mathcal{L}_s$, i.e., minimizing $\mathcal{L}$.

\begin{table*}[!t]
\centering
\caption{Quantitative comparison with state-of-the-art methods on the QaTa-COV19 dataset.}
\label{tab:comparison}
\begin{center}
\renewcommand\arraystretch{1.0}
\setlength{\tabcolsep}{7pt}
\begin{tabular}{lcc |lcc}
 \multicolumn{3}{c}{Uni-modal} & \multicolumn{3}{c}{Multi-modal}\\
\midrule[0.85pt]
Methods   & Dice & mIoU & Methods & Dice & mIoU\\
\midrule[0.5pt]
U-Net~\cite{ronneberger2015u}  &79.02  &69.46 &ConVIRT~\cite{zhang2022contrastive}   &79.72  &70.58\\
UNet++~\cite{zhou2018unet++}  &79.62   &70.25 &TGANet~\cite{tomar2022tganet}   &79.87  &70.75\\
AttUNet~\cite{oktay2018attention}  &79.31 &70.04 &GLoRIA~\cite{huang2021gloria}   &79.94  &70.68\\
nnUNet~\cite{isensee2021nnu}  &80.42  &70.81 &ViLT~\cite{kim2021vilt}  &79.63  &70.12\\
TransUnet~\cite{chen2021transunet}  &78.63  &69.13 &LAVT~\cite{yang2022lavt} &79.28  &69.89\\
Swin-Unet~\cite{cao2022swin}  &78.07  &68.34 &LViT~\cite{li2023lvit}  &83.66  &75.11\\
UCTransNet~\cite{wang2022uctransnet}  &79.15  &69.60 & CausalCLIPSeg & \textbf{85.21} & \textbf{76.90}\\

\bottomrule[0.85pt]
\end{tabular}
\end{center}
\label{tab:tab1}
\end{table*}

\section{Experiments}
\subsection{Dataset and Evaluation Metrics}
We conduct experiments on the QaTa-COV19 dataset, which consists of 9,258 COVID-19 chest X-ray radiographs with manual annotations of COVID-19 lesions. This dataset was compiled by researchers from Qatar University and Tampere University. A very recent work~\cite{li2023lvit} extends the QaTa-COV19 dataset with textual annotations, including whether both lungs are infected, the number of lesion areas, and the approximate location of the infected areas (e.g., ``Bilateral pulmonary infection, four infected areas, upper lower left lung and upper middle lower right lung''). The authors of~\cite{li2023lvit} partition the dataset into 5,716 training images, 1,429 validation images, and 2,113 test images.
\par
We adopt two commonly used evaluation metrics~\cite{everingham2015pascal}, Dice coefficient (Dice) and mean intersection-over-union (mIoU), to validate the effectiveness of the proposed and competing methods.

\subsection{Implementation Details}
We implement our approach using PyTorch and run it on a single NVIDIA RTX 4090 GPU. Input images are resized to $224\times224$ pixels. The maximum input sentence length is set to 20 tokens. We initialize the encoders with CLIP and use a ResNet-101 backbone for the vision encoder. We train for up to 2,000 epochs with an early stopping criteria after 100 epochs without validation improvement. Optimization uses Adam~\cite{kingma2014adam} with a cosine learning rate scheduler, starting from 3e-5, and an adversarial training coefficient \(\lambda\) of 0.05. Additionally, we conduct multiple experimental runs and observe minimal fluctuations in results. For instance, the standard deviation of Dice for our model does not exceed 0.002.

\begin{table*}[!t]
\centering
\caption{Ablation study of our proposed CausalCLIPSeg on the QaTa-COV19 dataset. We analyze the impact of different components of our approach.}
\label{tab:ablation}
\begin{center}
\renewcommand\arraystretch{0.95}
\setlength{\tabcolsep}{7pt}
\begin{tabular}{l cccc}
\midrule[0.85pt]
  & CLIP pre-trained & Causal intervention  & Dice & mIoU \\
\midrule[0.5pt]
1  &- &-  &82.50&73.24 \\
2  &- &\checkmark   &83.61&74.86 \\
3  &\checkmark &-   &83.71&74.69 \\
4  &\checkmark & \checkmark & \textbf{85.21} & \textbf{76.90}\\
\bottomrule[0.85pt]
\end{tabular}
\end{center}
\label{tab:tab1}
\end{table*}

\subsection{Comparison with State-of-the-Art Methods}
Table~\ref{tab:comparison} shows results on the QaTa-COV19 dataset, comparing the proposed CausalCLIPSeg to competitors. Our method outperforms the state-of-the-art LViT~\cite{li2023lvit} by 1.55\% Dice and 1.79\% mIoU. This demonstrates that our model effectively transfers the image-level knowledge of CLIP to pixel-wise segmentation tasks, having better cross-modal matching ability. Besides, our framework achieves remarkable performance gains of 5-7\% than vision-only methods~\cite{ronneberger2015u,zhou2018unet++,oktay2018attention,isensee2021nnu,chen2021transunet,cao2022swin,wang2022uctransnet}, highlighting the benefit of incorporating textual cues.
\par
\begin{figure}[t]
\begin{center}
\includegraphics[width = 0.95\linewidth]{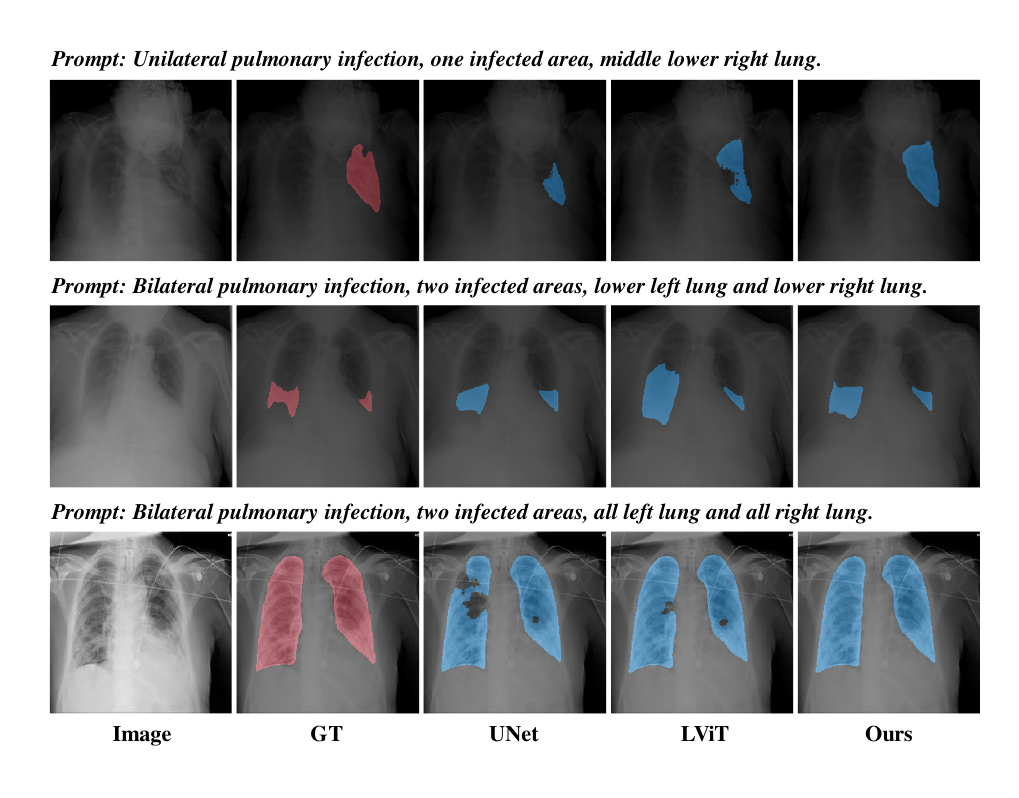}
\end{center}
\caption{Qualitative results on the QaTa-COV19 dataset.}
\label{comparison_fig}
\end{figure}

Fig.~\ref{comparison_fig} provides a qualitative comparison between CausalCLIPSeg and two representative methods, U-Net~\cite{ronneberger2015u} and LViT~\cite{li2023lvit}. It can be seen that our segmentations have less mis-labeled regions and more precise object boundaries. This shows our approach learns more robust semantic features and reduces confounding biases.

\subsection{Ablation Study}

We present ablation experiments in Table~\ref{tab:ablation} to evaluate the importance of several components. Note that adversarial masking is an indispensable component of our causal intervention module, and the adversarial loss is specifically designed to guide this module's learning. Therefore, we conduct an ablation study on the entire module. Specifically, we remove CLIP's pre-trained weights and the proposed causal intervention module from our framework to establish a baseline. As can be seen in the table, re-introducing both improves Dice by 2.71\% and mIoU by 3.66\%.
Ablating CLIP's pre-trained weights results in a 1.60\% decrease in Dice. In addition, removing the proposed causal intervention module also leads to a 1.50\% drop. Together, these ablation studies demonstrate the significance of both the CLIP encoders and causal intervention modeling for achieving state-of-the-art referring medical image segmentation.
\par
\begin{figure*}[!t]
\begin{center}
\includegraphics[width = 0.95\linewidth]{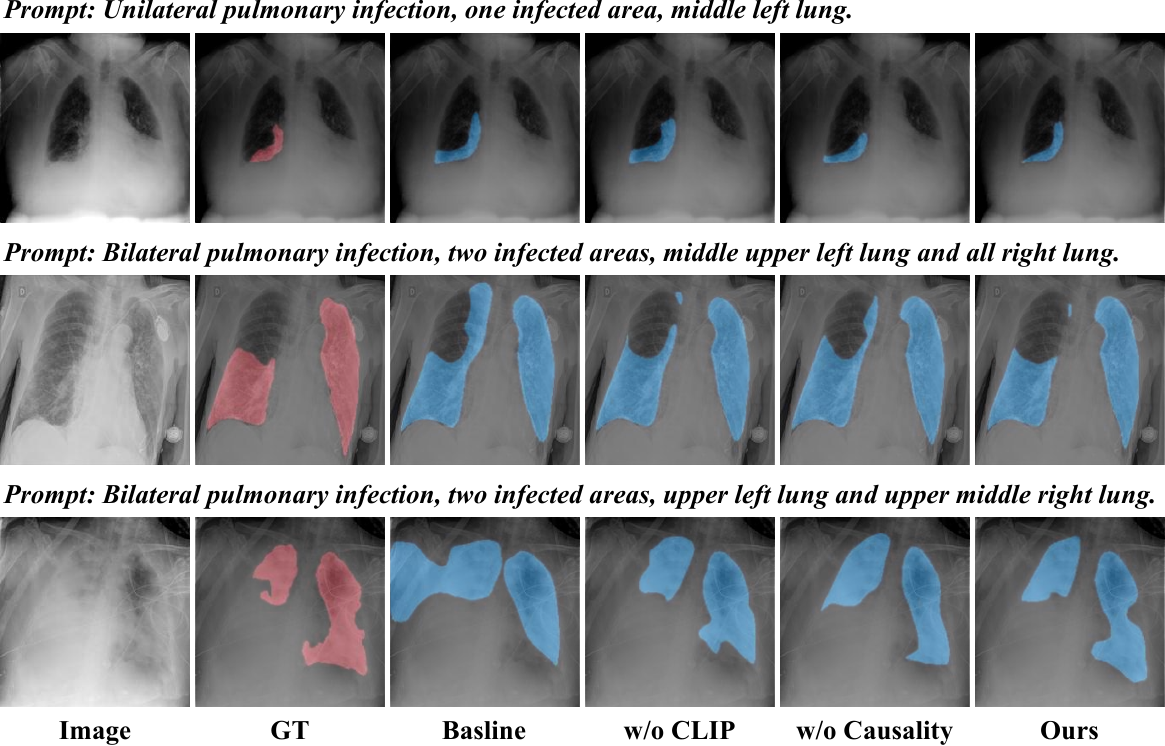}
\end{center}
\caption{Qualitative ablation study of CausalCLIPSeg.}
\label{ablation_fig}
\end{figure*}

Further, to gain more insights into the advantage of each component in our proposed approach, we show some visualization results in Fig.~\ref{ablation_fig}. Notably, the causal intervention module can alleviate poor segmentations when lesion and lesion-free regions are similar in appearance.

\section{Conclusion}
In this work, we propose CausalCLIPSeg, an end-to-end framework for referring medical image segmentation that leverages CLIP and a tailored cross-modal decoding method to align image pixels and textual cues. To mitigate confounding bias, we introduce a causal intervention module that self-annotates confounders and excavates causal features for more robust segmentation. An adversarial min-max game further optimizes causal features while penalizing confounding ones. Through extensive experiments, CausalCLIPSeg demonstrates state-of-the-art performance for referring medical image segmentation, showing strong potential to translate insights from large-scale pre-trained vision-language models to specialized medical applications. Future work includes extending our approach to other multi-modal medical data analysis tasks and investigating approaches to further improve robustness against spurious correlations.

\begin{credits}
\subsubsection{\ackname} This work is supported in part by the National Key Research and Development Program of China (2022ZD0160604), in part by the Natural Science Foundation of China (62101393/62176194), in part by the High-Performance Computing Platform of YZBSTCACC, and in part by MindSpore (\url{https://www.mindspore.cn}), a new deep learning framework.

\subsubsection{\discintname}
The authors have no competing interests to declare that are relevant to the content of this paper.
\end{credits}
%
%
%
%

\end{document}